Robert E. Kent
The Ontology Consortium, Pullman

# The Information Flow Foundation for Conceptual Knowledge Organization

**Abstract:** The sharing of ontologies between diverse communities of discourse allows them to compare their own information structures with that of other communities that share a common terminology and semantics – ontology sharing facilitates interoperability between online knowledge organizations. This paper demonstrates how ontology sharing is formalizable within the conceptual knowledge model of Information Flow (IF) (Barwise and Seligman, 1997). Information Flow indirectly represents sharing through a specifiable, ontology extension hierarchy augmented with synonymic type equivalencing – two ontologies share terminology and meaning through a common generic ontology that each extends. Using the paradigm of participant community ontologies formalized as IF logics, a common shared extensible ontology formalized as an IF theory, participant community specification links from the common ontology to the participating community ontology formalizable as IF theory interpretations, this paper argues that ontology sharing is concentrated in a virtual ontology of community connections, and demonstrates how this virtual ontology is computable as the fusion of the participant ontologies – the quotient of the sum of the participant ontologies modulo the ontological sharing structure.

## 1. Overview

Information Flow (IF), the logical design of distributed systems (Barwise and Seligman, 1997), provides a general theory of regularity that applies to the distributed information inherent in both the natural world of biological and physical systems and the artificial world of computational systems. According to (Barwise and Seligman, 1997) Information Flow "underlies the view of agents as information processors." In this paper we show how Information Flow provides a foundation for the sharing of ontologies in a distributed setting. In turn, the idea of ontology sharing illustrates the intuitions behind the Information Flow foundation for conceptual knowledge organization.

The sharing of ontologies between diverse communities of discourse allows them to compare their own information structures with that of other communities that share a common terminology and semantics. Ontologies and ontology sharing traditionally reside within the field of knowledge representation. Knowledge representation applies logic and ontology to knowledge organization. Logic supplies the form, consisting of framework and inferencing capabilities, whereas ontology supplies the content, consisting of the entities, relations, and constraints in the application domain.

The use of ontologies for knowledge organization is both old and new. In the far past, Aristotle effectively used ontological ideas in his system of classification. In the near past, Ranganathan effectively used dynamic ontological organizing principles (now understandable as ideas of conceptual knowledge organization) in his development of the Colon faceted classification system for organizing large research libraries. In the present and near future, various generic online knowledge organizing frameworks, such as RDF/S, OKBC, and OML/CKML (Kent, 1999), advocate the use of ontologies (schemas) for online activities such as e-commerce, bio-informatics, etc. The sharing of ontologies facilitates interoperability between online knowledge organizations (Kent, 1999).

Figure 1 illustrates the architecture of ontology sharing. By using the techniques and principles of Information Flow, this architecture can be constructed through a two-step process (Figure 3). It is the goal of this paper to explain how this process originates out of the foundations of Information Flow. The paper has five sections. Section 2 defines, reviews and introduces various concepts and properties of Information Flow. Here we discuss intuitions and give concrete definitions. Section 3 applies Information Flow to the conceptual organization of distributed knowledge, and presents the two-step process of ontology sharing. Section 4 summarizes and makes some observations about the study. The Notes section expands on the previous discussion in a more mathematical fashion, presenting new theorems on the representation and factorization of conceptual knowledge architecture.

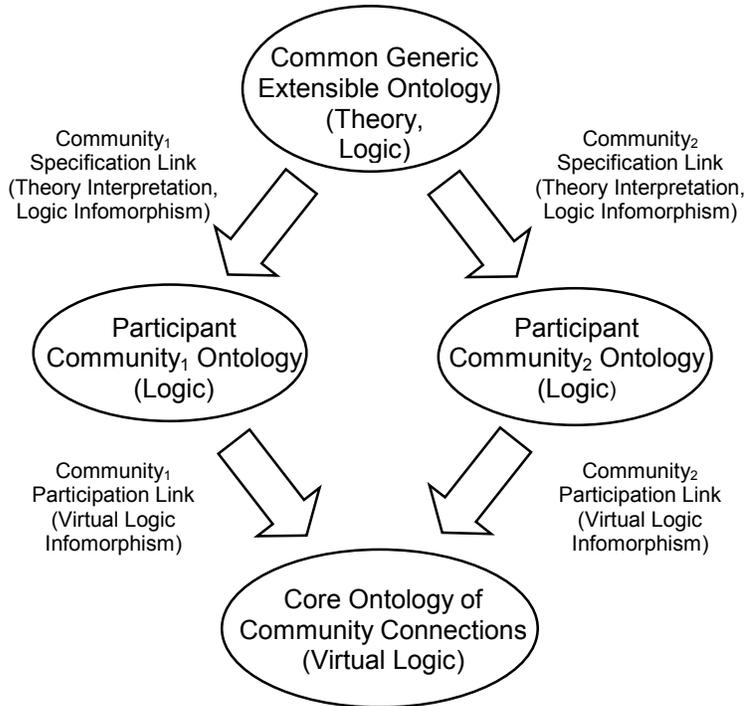

**Figure 1: Ontology Sharing between Communities**

## 2. The Foundations of Information Flow

According to the theory of information flow, information presupposes a system of classification. Classifications have been important in library science for the last 2,000 years. The library science classification system most in accord with the philosophy and techniques of Information Flow is the Colon classification system invented by the library scientist Ranganathan.

A domain-neutral notion of classification is given by the following abstract mathematical definition. A *classification* $A = \langle \mathit{inst}(A), \mathit{typ}(A), \vDash_A \rangle$ consists of:
1. a set $\mathit{inst}(A)$ of things to be classified, called the *instances* of $A$,
2. a set $\mathit{typ}(A)$ of things used to classify the instances, called the *types* of $A$, and
3. a binary relation, $\vDash_A$, from $\mathit{inst}(A)$ to $\mathit{typ}(A)$, called the *classification relation* of $A$.

The notation $a \vDash_A \alpha$ is read "instance $a$ is of type $\alpha$ in $A$." Classifications abound. Biologists classify organisms (instances) into categories (types). Linguists classify words (instances) by parts of speech (types). Classifications are related through infomorphisms. An *infomorphism* $\langle f, g \rangle : A_1 \rightleftarrows A_2$ from classification $A_1$ to classification $A_2$ consists of an instance function $f : \mathit{inst}(A_1) \leftarrow \mathit{inst}(A_2)$ and a type function $g : \mathit{typ}(A_1) \rightarrow \mathit{typ}(A_2)$ that satisfy the fundamental property, $f(a) \vDash_{A1} \alpha$ iff $a \vDash_{A2} g(\alpha)$, for $a \in \mathit{inst}(A_2)$ and $\alpha \in \mathit{typ}(A_1)$. To describe the architecture of Information Flow we use the intuitive terminology of contexts, passages and invertibility. The meaning of this terminology is defined in Table 1. Classifications and infomorphisms form the Classification context.

A *theory* is a pair $T = \langle typ(T), \vdash_T \rangle$, where $typ(T)$ is a set (of types) and $\vdash_T$ is a binary relation on $typ(T)$ called consequence. An element in the consequence relation, which is denoted as a sequent $\Gamma \vdash_T \Delta$ and is called a *constraint*, has the logical intention $\forall \Gamma \rightarrow \exists \Delta$ that "if all types in $\Gamma$ hold then some type in $\Delta$ holds." When $typ(A) = typ(T)$ for classification $A$, an instance $a \in inst(A)$ *satisfies* this constraint when it satisfies the intention: if instance $a$ is of every type in $\Gamma$, then it is of some type in $\Delta$. The theories generated by classifications (using satisfaction) obey structural axioms, such as *identity*, *weakening*, and *cut*. A (*theory*) *interpretation* $g : T_1 \rightarrow T_2$ is a function from $typ(T_1)$ to $typ(T_2)$ that preserves constraints: if $\Gamma_1 \vdash_{T1} \Delta_1$ then $g(\Gamma_1) \vdash_{T2} g(\Delta_1)$. Theories and interpretations form the Theory context. The Classification context and the Theory context are connected in opposite directions by invertible passages: the (theory) *Th* passage associates a theory with each classification, and the (classification) *Cla* passage associates a classification with each theory[1]. Using these invertible passages, the representation theorem of Information Flow (Barwise and Seligman, 1997) has a generalization that is suitable for distributed conceptual knowledge organization[2].

| INTUITIVE TERMINOLOGY | MATHEMATICAL MEANING |
|---|---|
| context | category |
| passage | functor |
| invertible | adjoint |
| sum | coproduct |
| quotient | — |
| fusion | pushout |

**Table 1: Definition of Terminology**

A *local logic* $L = \langle inst(L), typ(L), \vDash_L, \vdash_L, N_L \rangle$, which is an inclusive idea combining the notions of classification and theory into a (not necessarily sound) whole, consists of

1. a theory $th(L) = \langle typ(L), \vdash_L \rangle$ of types and constraints,
2. a classification $cla(L) = \langle inst(L), typ(L), \vDash_L \rangle$ of instances,
3. a subset $N_L \subseteq inst(L)$ of *normal instances* which satisfy all the constraints.

A logic is *sound* when every instance of $inst(L)$ is normal. For any local logic $L$, the *sound part* of $L$ is obtained by throwing away all abnormal instances and restricting the classification relation to normal instances. In this paper we limit ourselves to sound logics, since these enable ontology sharing (see below).

A (*sound*) *logic* $L = \langle inst(L), typ(L), \vDash_L, \vdash_L \rangle$ consists of

1. a theory $th(L) = \langle typ(L), \vdash_L \rangle$ of types and constraints, and
2. a classification $cla(L) = \langle inst(L), typ(L), \vDash_L \rangle$ of instances which satisfy all the constraints.

A (*logic*) *infomorphism* $\langle f, g \rangle : L_1 \rightleftharpoons L_2$ is both an classification infomorphism $\langle f, g \rangle : cla(L_1) \rightleftharpoons cla(L_2)$ and a theory interpretation $g : th(L_1) \rightarrow th(L_2)$. Logics and infomorphisms form the Logic context. There is an underlying classification passage $cla : \text{Logic} \rightarrow \text{Classification}$. There is also an underlying theory passage $th : \text{Logic} \rightarrow \text{Theory}$. Both of these have inverse "free logics" passages[3]. Figure 2, which illustrates the contexts and invertible passages involved in the architecture of Conceptual Knowledge Organization, is the factorization through the Logic context of the generalized representation theorem invertibility[4]. The invertible passages *Log* and *th*, which are illustrated in bold on the right side of Figure 2, are central to this paper. They connect the dynamism of the Logic context with the stability of the Theory context.

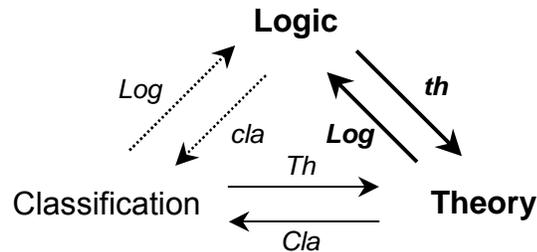

**Figure 2: Conceptual Knowledge Organization**

## 3. The Representation of Ontological Structures

Information Flow represents both the dynamism and stability of conceptual knowledge organization. Stability is represented by the types and constraints specified within ontologies and formalized within the `Theory` context as the theories and theory interpretations of Information Flow. Dynamism is represented by instance collections, their classification relations, and the links between ontologies specified by ontological extension and synonymy (type equivalence) and formalized within the `Logic` context as the logics and logic infomorphisms of Information Flow. Ontologies and ontology sharing, which are formalizable within Information Flow and its conceptual knowledge model, manifest this dynamism and stability. Information Flow represents ontologies as logics. It represents ontology sharing through a specifiable ontology extension hierarchy. An ontology has a classification relation between instances and types and a set of constraints modeling the ontology's semantics. These constraints can represent subtyping, partition, disjointness, covering and incoherence. Since ontologies exist in the distributed setting they can be of varying quality, and are best represented by the (possibly unsound) local logic of Information Flow. However, the logics needed in ontology sharing must all be sound. So, in order to use ontologies in ontology sharing, we must work with their sound part – ontology sharing requires quality information.

The terminology and semantics of a community's knowledge is specified in an ontology, and realized within the various instance collections of that community. A *community ontology* (Figure 1) is the basic unit of ontology sharing. Community ontologies share terminology and constraints through a common generic ontology that each extends. In ontology sharing, the constraints in participant community ontologies represent consensual agreement *within* those communities, whereas the constraints in generic common extensible ontologies represent consensual agreement *across* communities – a standard semantics. A *common generic extensible ontology* (Figure 1) consists of the common terminology and semantics shared by diverse communities. This is formalized as a theory with neither *a priori* instances nor classification relation. Formal instances can be added in the passage to (construction of) a "free" logic. The actual accessible instances of the generic ontology can either be incorporated within the instances of any specific ontology or can be ignored. *Specification links* connect the common ontology with participating community ontologies. These links represent ontological extension: they *include* the types and constraints of the common ontology, and they record any synonymy or *type equivalence* prescribed by the participants. To enable ontology sharing in a distributed environment, several principles are being used.

***Principle 1.*** A community owns its collection of instances: (a) it controls updates to the collection; (b) it can enforce soundness; (c) it controls access rights to the collection.

***Principle 2.*** Instances are linked through their types – in order to be able to compare instances of two specific ontologies, we must use the free logic of the generic ontology containing all of its formal instances.

There is a natural set of "connections" between instances of participating communities: an instance of one community is connected to an instance of another community when they agree on the common inherited types. Because they are combinations of instances from the classifications for the participating communities, to classify such connections we can use the sum[5] of the types in the community participant ontologies. However, we must identify types that are linked to each other by a common ontology type through the specification links. Instance connections and identified types comprise a natural quotient construction[6] on the participating community ontologies. The *virtual ontology of community connections* (Figure 1) is computable as this quotient. This virtual ontology is a fusion of the participating ontologies[7]: it represents the complete system of ontology sharing, and the participating community ontologies can be recovered as projections.

Figure 3 represents the process of ontology sharing. Here the virtual ontology of community connections is computed. As illustrated in Figure 3, the process of ontology sharing consists of two steps connecting three sharing diagrams. The first two diagrams represent the *specification* of ontology sharing, whereas the third fusion diagram represents the *computation* of ontology sharing, which results in the ontology of community connections. The first specification diagram exists within the Theory context, whereas the second specification diagram and the fusion diagram exist within the Logic context. $L_1/E_1$ and $L_2/E_2$ are the participant community logics with synonymic type equivalences $E_1$ and $E_2$. $Log(T)$ is the free logic for the common generic extensible ontology. $L_1/E_1 +_{Log(T)} L_2/E_2$ is the virtual logic of community connections, which is the fusion of community logics via type inclusion with synonymy (type equivalence). $g_1 : T \to typ(L_1)$ and $g_2 : T \to typ(L_2)$ represent the community specification links in the Theory context, and $\langle f_1, g_1 \rangle : Log(T) \rightleftharpoons L_1$ and $\langle f_2, g_2 \rangle : Log(T) \rightleftharpoons L_2$ represent the community specification links in the Logic context. Finally, $\langle \tilde{f}_1, \tilde{g}_1 \rangle$ and $\langle \tilde{f}_2, \tilde{g}_2 \rangle$ are the component embeddings into the fusion logic.

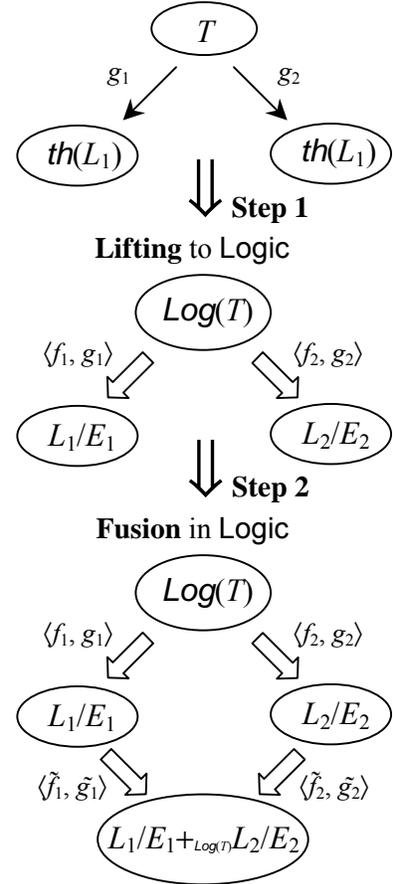

**Figure 3: Two-step Process**

The purpose of the first step in the ontology sharing process is the conversion of specification between the two contexts. This step lifts the generic common extensible ontology from the Theory context to the Logic context by adding all of its formal instances. It also transforms between the adjoint forms of the specification links by using the free logic passage *Log* and its invertibility with the underlying theory passage *th*. For this to be possible the participating community ontologies are required to be sound – they are required to satisfy all constraints. The purpose of the second step in the ontology sharing process is construction of the virtual ontology of community connections. Operating within the Logic context, it starts with the second specification diagram and ends with the third fusion diagram, an abstraction of Figure 1. First, compute the sum of the participant ontologies. Second, form the quotient logic specified by linkage invariance through the generic logic. Since the specification links represent both type inclusion and synonymy, this will consist of interleaving synonymy with sharing linkage. Soundness is preserve by fusion. Since soundness is required in the first step, quality information is a prerequisite for ontology sharing in a distributed environment. Since soundness is preserved in the second step, quality information is assured overall.

## 4. Summary

Ontology sharing can be successfully founded upon principles of Information Flow. The logics of Information Flow are adequate from the representational standpoint: ontologies are best described using local logics, and sound logics are needed for the ontology sharing described here. Communities are the unit of interoperability, not individuals. Sharing interoperability occurs through generic ontologies viewed as theories (types and constraints) with no instance collections. If they exist and a community deems them of interest, generic instance collections can be included as part of the specific instance collection of the community.

# Notes

[1] Any classification $A$ has an associated theory $Th(A) = \langle typ(A), \vdash_A \rangle$, whose types are types of $A$ and whose constraints are the set of sequents satisfied by every instance in $A$. $Th(A)$ satisfies the axioms of identity, weakening, cut and *partition*: If $\Gamma' \vdash \Delta'$ for each partition of $typ(T)$ with $\langle \Gamma, \Delta \rangle \leq \langle \Gamma', \Delta' \rangle$ then $\Gamma \vdash \Delta$. Any theory that satisfies these four axioms is called a *regular theory*. $Th(A)$ is regular. Any infomorphism $\langle f, g \rangle : A \rightleftarrows B$ has an associated theory interpretation $Th(\langle f, g \rangle) = g : Th(A) \to Th(B)$. These associations define the *theory functor* $Th$ : Classification → Theory, which factors through the subcategory of regular theories Regular Theory.

Any theory $T$ has an associated classification $Cla(T) = \langle inst(T), typ(T), \vDash_T \rangle$, whose instances are the formal instances of $T$, whose types are the types of $T$, and whose classification $\vDash_T$ is defined by: $\langle \Gamma, \Delta \rangle \vDash_T \alpha$ iff $\alpha \in \Gamma$ (or equivalently $\alpha \notin \Delta$). Any theory interpretation $g : T_1 \to T_2$ has an associated infomorphism $Cla(g) = \langle g^{-1}, g \rangle : Cla(T_1) \rightleftarrows Cla(T_2)$, whose type function is $g$ itself, and whose instance function is inverse image defined by $g^{-1}(\langle \Gamma_2, \Delta_2 \rangle) = \langle g^{-1}(\Gamma_2), g^{-1}(\Delta_2) \rangle$. These associations define the *classification functor* $Cla$ : Theory → Classification.

[2] The following theorem is an adjointness generalization of the representation theorem of (Barwise and Seligman, 1997).

> **Representation Theorem:** *The classification functor $Cla$ is left adjoint to the theory functor $Th$, $Cla \dashv Th$. When restricted to the subcategory of regular theories, the counit natural transformation $\varepsilon$ of the adjunction becomes an identity and the classification functor embeds the category of regular theories into the category of classifications:* Classification ⊃ Regular Theory. *In particular, for any regular theory $T = Th(Cla(T))$ and for any regular theory interpretation $g = Th(Cla(g))$. When restricted to both the subcategory of separated classifications and the subcategory of regular theories, the adjunction becomes an equivalence of categories*: Separated Classification ≡ Regular Theory.

The unit of the adjunction $Cla \dashv Th$ is a natural transformation $\eta : Id_{\text{Theory}} \Rightarrow Cla \cdot Th$, whose $T$-th component is the regular closure inclusion theory interpretation $\eta_T : T \to Th(Cla(T))$. The counit is the natural transformation $\varepsilon : Th \cdot Cla \Rightarrow Id_{\text{Classification}}$, whose $A$-th component is the infomorphism $\varepsilon_A = \langle Id_{typ(A)}, \sigma_A \rangle : Cla(Th(A)) \rightleftarrows A$, which is the identity on types and the state description on instances.

[3] Any classification $A$ has an associated logic $Log(A)$, whose classification is $A$ itself, and whose (regular) theory is $Th(A)$. Any classification infomorphism $\langle f, g \rangle : A \rightleftarrows B$ has an associated logic infomorphism $Log(\langle f, g \rangle) : Log(A) \rightleftarrows Log(B)$, which is the same contravariant pair of functions. These associations define the "free" *logic* functor $Log$ : Classification → Logic.

Any theory $T$ has an associated logic $Log(T)$, whose classification is $Cla(T)$, and whose theory is $T$ itself. Any theory interpretation $g : T_1 \to T_2$ has an associated logic infomorphism $Log(g) : Log(T_1) \rightleftarrows Log(T_2)$, whose type function is $g$ itself, and whose instance function is the inverse image on formal instances $g^{-1}(\langle \Gamma, \Delta \rangle) = \langle g^{-1}(\Gamma), g^{-1}(\Delta) \rangle$. These associations define the "free" *logic* functor $Log$ : Theory → Logic (add formal instances).

[4] The following theorem factors the adjointness generalization of representation.

> **Factorization Theorem:** *The $Cla \dashv Th$ adjunction factors through the category of sound logics. The logic functor $Log$ on theories is left adjoint to the underlying theory functor $th$ on sound logics, $Log \dashv th$. The underlying classification functor $cla$ on sound logics is left adjoint to the logic functor $Log$ on classifications, $cla \dashv Log$. The $Cla \dashv Th$ adjunction factors as the composition of these two adjunctions,* $(Cla \dashv Th) = (cla \dashv Log) \circ (Log \dashv th)$.

The unit of the adjunction $Log \dashv th$ is the identity natural transformation $\eta : Id_{\text{Theory}} = Log \cdot th$. The counit is the natural transformation $\varepsilon : th \cdot Log \Rightarrow Id_{\text{Logic}}$, whose $L$-th component is the infomorphism $\varepsilon_L = \langle Id, \sigma_L \rangle : Log(th(L)) \rightleftarrows L$, whose type function is identity, and whose instance function is state description $\sigma_L : inst(L) \to inst(th(L))$ mapping any instance $a \in inst(L)$ to the formal instance $\sigma_L(a) = \langle typ(a), typ(L) - typ(a) \rangle$. The $Log \dashv th$ adjunction between theories and sound logics, is the adjunction used to define the sharing of ontologies. In one sense we can regard the adjunction factorization to be the lifting of the $Cla \dashv Th$ adjunction to the $(Log \dashv th)$ adjunction along the $(cla \dashv Log)$ adjunction.

⁵ Given two classifications $A_0$ and $A_1$, the *sum* $A_0+A_1$ is the classification, whose instance set is the has Cartesian product $\textsf{inst}(A_0+A_1) = \textsf{inst}(A_0) \times \textsf{inst}(A_1)$, whose type set is the disjoint union $\textsf{typ}(A_0+A_1) = \textsf{typ}(A_0)+\textsf{typ}(A_1)$, and whose classification relation is defined by either $(a_0, a_1) \vDash_{A0+A1} \alpha_0$ iff $a_0 \vDash_{A0} \alpha_0$ or $(a_0, a_1) \vDash_{A0+A1} \alpha_1$ iff $a_1 \vDash_{A1} \alpha_1$. Each sum classification comes equipped with two mediating infomorphisms $\iota_0 = \langle pr_0, in_0 \rangle : A_0 \rightleftarrows A_0+A_1$ and $\iota_1 = \langle pr_1, in_1 \rangle : A_1 \rightleftarrows A_0+A_1$ defined as injection on types and projection on instances. With these infomorphisms the classification sum is a coproduct in Classification.

Given two theories $T_0$ and $T_1$, the *sum* $T_0+T_1$ is the theory, whose type set is the disjoint union $\textsf{typ}(T_0+T_1) = \textsf{typ}(T_0)+\textsf{typ}(T_1)$, and whose consequence relation is the union of the component consequence relations: $\Gamma \vdash_{T0+T1} \Delta$ iff $\Gamma \vdash_{T0} \Delta$ or $\Gamma \vdash_{T0} \Delta$. Each sum theory comes equipped with two mediating interpretations $in_0 : T_0 \to T_0+T_1$ and $in_1 : T_1 \to T_0+T_1$ defined as injection on types. With these interpretations the theory sum is a coproduct in Theory.

Given two (sound) logics $L_0$ and $L_1$, the *sum* $L_0+L_1$ is the logic, whose classification is the sum classification $\textsf{cla}(L_0+L_1) = \textsf{cla}(L_0)+\textsf{cla}(L_1)$, and whose theory is the sum theory $\textsf{th}(L_0+L_1) = \textsf{th}(L_0)+\textsf{th}(L_1)$. Each logic sum comes equipped with two mediating logic infomorphisms $\iota_0 : L_0 \rightleftarrows L_0+L_1$ and $\iota_1 : L_1 \rightleftarrows L_0+L_1$ defined to be the same as the classification infomorphisms $\iota_k : \textsf{cla}(L_k) \rightleftarrows \textsf{cla}(L_0)+\textsf{cla}(L_1)$ for $k = 0,1$. Then the type functions are also theory interpretations $in_k : \textsf{th}(L_k) \to \textsf{th}(L_0)+\textsf{th}(L_1)$ for $k = 0,1$. With these infomorphisms the logic sum is a coproduct in Logic.

⁶ Given a classification $A$, a *dual invariant* is a pair $J = (X, R)$ consisting of a subset $X \subseteq \textsf{inst}(A)$ of instances of $A$ and a binary relation $R$ on types of $A$ satisfying the constraint: if $\alpha R \beta$, then for each $a \in X$, $a \vDash_A \alpha$ iff $a \vDash_A \beta$. The *dual quotient* of $A$ by $J$ is the classification $A/J$, whose instance set is $X$, whose types are the $R$-equivalence classes of types of $A$, and whose classification relation is: $a \vDash_{A/J} [\alpha]$ iff $a \vDash_A \alpha$. The pair $\tau_J = \langle \textsf{incl}_A, [\ ]_J \rangle : A \rightleftarrows A/J$, whose instance function is the inclusion of $A$ and whose type function is the canonical map $[\ ]_J : \textsf{typ}(A) \to \textsf{typ}(A/J)$, is a mediating infomorphism. If $\langle f, g \rangle : A \rightleftarrows A'$ is any classification infomorphism that respects $J$, in the sense that $f(a') \in X$ for any instance of $A'$ and $g(\alpha) = g(\beta)$ whenever $\alpha R \beta$, then there is a unique classification infomorphism $\langle f', g' \rangle : A/J \rightleftarrows A'$ such that $\langle f, g \rangle = \tau_J \circ \langle f', g' \rangle$.

For any a binary relation $R$ on a set $\Sigma$ and any subset $\Gamma \subseteq \Sigma$, the *equivalence image* of $\Gamma$ with respect to $R$ is the set of $R$-equivalence classes covering $\Gamma$: $[\Gamma]_R = \{[\alpha]_R \mid \alpha \in \Gamma\}$. Given a theory $T = \langle \textsf{typ}(T), \vdash_T \rangle$, a *dual invariant* is any binary relation $R$ on types of $T$. The *dual quotient* of $T$ by $R$, written $T/R$, is the theory whose types are the $R$-equivalence classes of types of $T$, and whose consequence relation consists of all "equivalence images" $[\Gamma]_R \vdash_{T/R} [\Delta]_R$. The canonical map $[\ ]_R : T \to T/R$ is a theory interpretation. If $g : T \to T'$ is any theory interpretation that respects $R$, in the sense that $g(\alpha) = g(\beta)$ whenever $\alpha R \beta$, then there is a unique theory interpretation $g' : T/R \to T'$ such that $g(\alpha) = g'([\beta]_R)$.

Given a (sound) logic **L**, a *dual invariant* is a pair $J = (A, R)$ that is a dual invariant on both $\textsf{cla}(J)$ and $\textsf{th}(L)$. The *dual quotient* of $L$ by $J$, written $L/J$, is the logic whose classification is $\textsf{cla}(L)/J$ and whose theory is $\textsf{th}(L)/R$. This is well-defined, since any instance $a \in A$ satisfies any constraint $[\Gamma]_R \vdash_{T/R} [\Delta]_R$.

⁷ For any two logic infomorphisms $\langle f_0, g_0 \rangle : L \rightleftarrows L_0$ and $\langle f_1, g_1 \rangle : L \rightleftarrows L_1$ that share a common source logic $L$, we define a dual invariant $J = (A, R)$ on the sum $L_0+L_1$. Let $A$ be the pairs of instances $(a_0, a_1)$, $a_0 \in \textsf{inst}(L_0)$ and $a_1 \in \textsf{inst}(L_1)$, on which $f_0$ and $f_1$ agree, $f_0(a_0) = f_1(a_1)$, and let $R$ be the binary relation between types of $L_0$ and types of $L_1$ defined by $\alpha_0 R \alpha_1$ when there is a $\alpha \in \textsf{typ}(L)$ such that $g_0(\alpha) = \alpha_0$ and $g_1(\alpha) = \alpha_1$. The *fusion* logic is the dual quotient $L_0 +_L L_1 = (L_0+L_1)/J$. The logic infomorphism $\tau_J : L_0+L_1 \rightleftarrows L_0+_L L_1$ defines by composition two mediating infomorphisms $\langle \tilde{f}_0, \tilde{g}_0 \rangle = \iota_0 \circ \tau_J : L_0 \rightleftarrows L_0+_L L_1$ and $\langle \tilde{f}_1, \tilde{g}_1 \rangle = \iota_1 \circ \tau_J : L_1 \rightleftarrows L_0+_L L_1$. With these infomorphisms the fusion logic is a pushout in Logic.